\documentclass[11pt]{article}

\usepackage[final]{acl}

\usepackage{times}
\usepackage{latexsym}

\usepackage[T1]{fontenc}
\usepackage[utf8]{inputenc}

\usepackage{microtype}

\usepackage{inconsolata}

\usepackage{graphicx}
\usepackage{amsmath}
\usepackage{bbm}

\usepackage{tikz}
\usetikzlibrary{arrows.meta, positioning, decorations.pathreplacing, fit, backgrounds, calc}

\title{Less Is More: Graph-free Multimodal RAG \\ via Multi-signal Late Fusion}

\author{
  Tithi Rakshit$^{1}$\thanks{~Work done during an internship at Robert Bosch GmbH.} \quad Hongkuan Zhou$^{2,3}$ \quad Lavdim Halilaj$^{2}$ \quad Yuqicheng Zhu$^{2,3}$ \\
  $^{1}$University of T\"ubingen, T\"ubingen, Germany \\
  $^{2}$Corporate Research, Robert Bosch GmbH, Renningen, Germany \\
  $^{3}$University of Stuttgart, Stuttgart, Germany \\
  \texttt{tithi.rakshit@student.uni-tuebingen.de} \\
  \texttt{\{hongkuan.zhou, lavdim.halilaj, yuqicheng.zhu\}@de.bosch.com}
}

\begin{document}
\maketitle
\begin{abstract}
Graph-based retrieval-augmented generation (RAG) is widely used for multimodal, cross-document question answering. However, building corpus-level graphs is expensive, slow to query, and difficult to maintain.
We present TrioRAG, a graph-free multimodal framework that integrates evidence from three complementary signals: the question, the anchor image, and a VLM-enhanced query generated from both.
Each signal retrieves independently over a shared multi-vector index of page text and page images, and the results are combined through late fusion.
Further, we introduce AutoQA, a multimodal automotive benchmark whose questions are grounded in noisy, web-sourced images rather than clean document-sourced figures. Its questions require reasoning across manuals. 
We position it as a model-curated testbed rather than a human-validated gold standard. 
Across three benchmarks, TrioRAG matches or outperforms graph-based systems while reducing total cost and accelerating per-query inference by 1.6-2.3$\times$.
By construction, AutoQA grounds its questions in out-of-corpus web images. In this setting image retrieval reaches only 19.3\% document-level recall, while text-derived signals, especially the VLM-enhanced query, keep retrieval robust.
\end{abstract}

\section{Introduction}
Retrieval-Augmented Generation (RAG) \citep{lewis2020retrieval} has become the standard paradigm for grounding large language models in external knowledge. For multimodal and cross-document reasoning, recent systems increasingly turn to graph-based designs \citep{edge2024local, guo2024lightrag, wan2025mmgraphrag, guo2025rag} that build graphs over documents, pages, entities, text, and visual elements. Such a structure is a natural fit for multimodal corpora, where evidence is spread across text, figures, tables, and images.

\begin{figure}[t] 
\centering
\includegraphics[
    width=\columnwidth,
    trim=0.7cm 0.3cm 0.7cm 0.3cm,
    clip
]{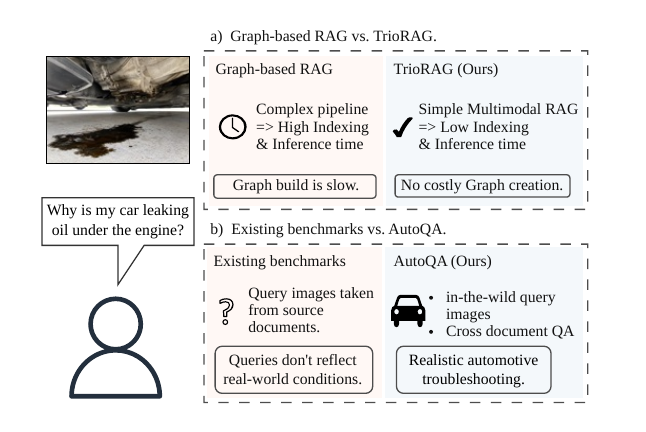}
\caption{
\textbf{Vehicle Diagnostics: RAG Pipelines and Benchmarks}: 
a) \textit{Overhead-free}: TrioRAG achieves strong performance without graph construction or complex retrieval pipelines.
b) \textit{Out-of-corpus images}: AutoQA uses messy, web-sourced diagnostic images rather than images drawn from the source corpus.}
  \label{fig:motivation}
\end{figure}

However, this expressivity is costly (Figure~\ref{fig:motivation}). Graph-based systems require expensive extraction, alignment, and construction, and are difficult to update, as corpus changes can force re-extraction, or partial rebuilds \citep{zhang2026respecting, peng2025graph}. These challenges are amplified in multimodal settings, where text, images, and entities must be linked across large collections.
This raises a practical question for the graph systems and benchmarks we study: is a corpus-level graph necessary for competitive multimodal cross-document retrieval, or can simpler retrieval methods extract comparable evidence at lower cost?

To address this challenge, we propose TrioRAG, a graph-free multimodal RAG framework.\footnote{Code: \url{https://github.com/hk-zh/TrioRAG}} TrioRAG indexes page text and images in a shared multi-vector space and retrieves three complementary signals: the text query, the anchor image, and a VLM-enhanced query grounded on the anchor image. The resulting rankings are fused via late fusion. By avoiding graph construction and traversal, TrioRAG achieves performance comparable to or better than the graph-based systems we evaluate.

Existing multimodal RAG benchmarks usually draw query images from source documents, creating near-duplicate visual matches that can overestimate the effectiveness of image retrieval compared to out-of-corpus query images, which are common in real-world settings.
To evaluate retrieval under realistic conditions, we introduce AutoQA, a multimodal, cross-document question-answering benchmark in the automotive domain. 
Each question is grounded in a noisy, web-sourced image and requires synthesizing evidence across two technical manuals. AutoQA deliberately creates a domain gap between query images and document pages, allowing us to study retrieval when no direct visual match exists in the corpus. 
Our experiments reveal that, in this setting, image retrieval becomes substantially less reliable, while text-derived signals, especially the VLM-enhanced query, recover much of the missing evidence and keep retrieval robust. 

Our contributions are as follows: \textbf{A graph-free framework} that fuses three query signals: the text query, the anchor image query, and a VLM-enhanced query, over a shared multi-vector index, matching or exceeding the graph-based systems we evaluate.
\textbf{AutoQA}, a model-curated benchmark of multimodal, cross-document automotive QA pairs grounded in noisy, web-sourced images, with validation status discussed in the Limitations and Ethics statements.
\textbf{A retrieval-reliability analysis} showing that, on AutoQA's deliberately out-of-corpus query images, image retrieval collapses while text-derived signals, especially the VLM-enhanced query, keep retrieval robust. On the benchmarks we study, this reliability gap, rather than the choice of fusion method, is what most closely tracks downstream accuracy.

\section{Related Work}

\textbf{Retrieval-augmented generation (RAG)} \citep{lewis2020retrieval} grounds language models in external knowledge. It does so by retrieving relevant passages with a dense retriever \citep{karpukhin2020dense} and conditioning a reader on the retrieved evidence. Fusion-in-Decoder \citep{izacard-grave-2021-leveraging} and Atlas \citep{izacard2023atlas} show that jointly attending to multiple passages improves performance, motivating multi-evidence readers rather than single-match retrieval. However, these methods assume textual knowledge, limiting their use for visual documents. OCR or captioning pipelines \citep{xu2020layoutlm, riedler2024beyond} are often used instead, but they flatten layouts and discard visual structure.

\textbf{Visual document retrieval.} Recent work retrieves directly from visual inputs. Building on late interaction models \citep{khattab2020colbert}, ColPali \citep{faysse2025colpali} uses multi-vector page embeddings with MaxSim, avoiding OCR. VisRAG \citep{yu2025visrag} and M3DocRAG \citep{cho2025m3docvqa} show strong performance in document QA at scale, and ViDoRAG \citep{wang2025vidorag} adds iterative reasoning over retrieved pages. We use multi-vector retrieval but differ in query construction. We combine multiple complementary query signals instead of a single representation.

\textbf{Query expansion for retrieval.} Query expansion methods generate intermediate text to improve retrieval. HyDE \citep{gao2023precise} prompts a language model to generate a hypothetical document, whose embedding is used for retrieval. We extend this to multimodal inputs by conditioning generation on both the question and an anchor image, grounding abstract terms in visual content. This serves as one of several retrieval signals rather than a standalone method.

\textbf{Graph-based RAG.} RAG systems increasingly use graph structures to support retrieval and reasoning \citep{guo2024lightrag, zhou2025evaluating, zhou2025gragent, zhou2026breaks, chaturvedi-etal-2026-scair}. In multimodal settings, MMGraphRAG \citep{wan2025mmgraphrag} aligns image and text graphs via spectral clustering for entity linking, while RAG-Anything \citep{guo2025rag} extends graph-based retrieval to text and vision. VAT-KG \citep{park2025vat} constructs concept-level graphs across modalities. Graph structure can also be introduced after retrieval for structured reasoning, e.g., argumentation-based RAG \citep{pmlr-v284-zhu25a}. However, methods that construct corpus-level graphs can incur substantial costs from extraction, alignment, and traversal.

\textbf{Rank aggregation and fusion.} Rank aggregation has a long history in information retrieval \citep{li2025bordarag, santra2025hf, rackauckas2024rag}. Classical methods like Borda count \citep{aslam2001models} and Reciprocal Rank Fusion \citep{cormack2009reciprocal} combine sparse and dense signals. 
Social-choice aggregation has also been used to robustly combine LLM outputs \citep{robustllm}.
Recent work extends them to multimodal settings \citep{samuel2025mmmorrf}. We use unsupervised fusion methods and study how they behave under heterogeneous signal reliability, especially with noisy query images.

\textbf{Benchmarks.} Document QA benchmarks evolve from single-page \citep{mathew2021docvqa} to multi-page \citep{cho2025m3docvqa} and long-document reasoning \citep{ma2024mmlongbench, li2024m3sciqa}. However, most assume clean inputs and aligned visual evidence, limiting realism. To address this gap, AutoQA introduces cross-document automotive reasoning with web-sourced question images.

\section{Methodology}
\label{sec:methodology}
\begin{figure*}[t]
  \centering
  \includegraphics[width=1.0\textwidth]{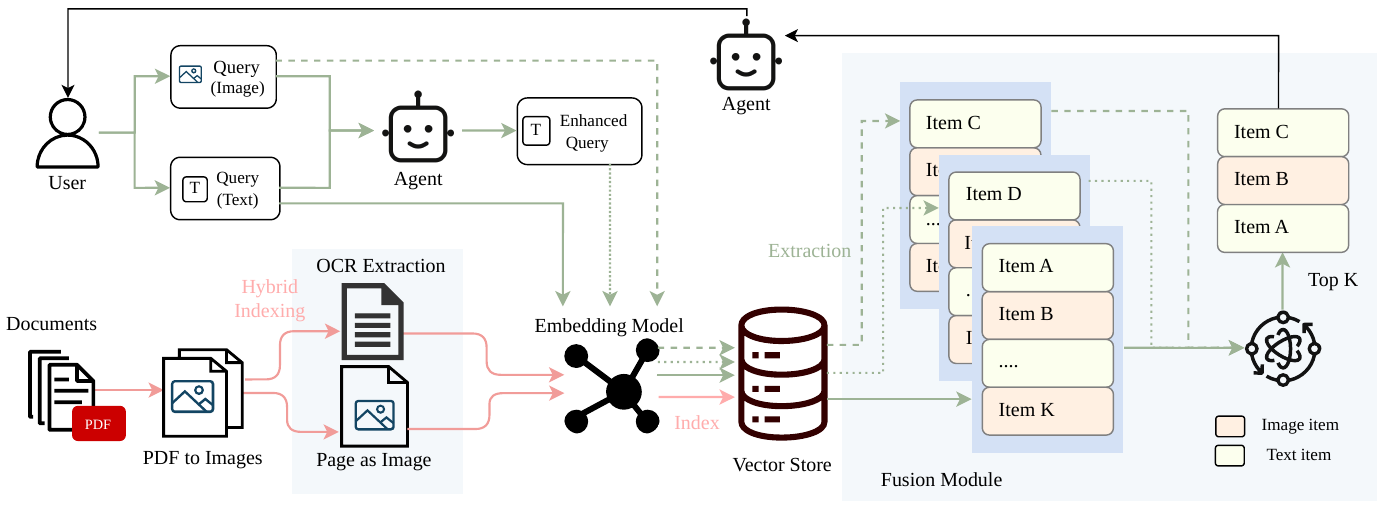}
  \caption{Our TrioRAG pipeline. The multimodal question is turned into three signals: text, image, and VLM-enhanced text, retrieved independently over a hybrid multi-vector index and merged into a single fused ranking. The top-ranked pages are passed to the VLM reader alongside the anchor image to generate the answer.}
  \label{fig:architecture}
\end{figure*}

Given a question $q$, an anchor image $v$, and a corpus of $M$ pages $\mathcal{P}=\{p_j\}_{j=1}^{M}$, our pipeline retrieves a set of pages $\mathcal{R}_K\subset\mathcal{P}$ and passes them, together with $(q,v)$, to a vision-language reader that produces the answer $a$. The pipeline has three stages: hybrid indexing, multi-signal retrieval, and late fusion, followed by a generation step, with no graph construction at any point (Figure~\ref{fig:architecture}). We describe the default multi-vector configuration here; a dense variant is evaluated as an ablation (Section~\ref{sec:abl-indexing}).

\textbf{Hybrid indexing.} Each page $p_j$ is stored under two views: its structured text $x_j$, extracted with MinerU \citep{wang2024mineru}, and its rendered page image $y_j$. Both are encoded with Jina-v4 \citep{gunther2025jina},
a unified multimodal encoder that maps text and images into a shared embedding space, and they are added to a single index. Keeping a text and an image view of every page preserves both textual content and visual layout without any structural parsing, enabling the query match of either view.

\textbf{Multi-signal queries.} We derive three retrieval signals from the multimodal question $(q,v)$, each surfacing different evidence. 
A text query ($q$), an image query ($v$), and an enhanced query $e$ i.e., a short textual description produced by prompting a VLM with both $q$ and $v$ (prompt in Appendix~\ref{app:prompts}). 
The enhanced question grounds abstract question terms in what is visible in the anchor image and is embedded as text. 
Combining the three signals improves recall over any single one (Table~\ref{tab:retrieval_regime}).

\textbf{Late-interaction retrieval.} 
Each question is a set of token embeddings. For a question $Q$ and an index entry $c$ with embedding set $V_c$, we score relevance by late interaction (MaxSim) \citep{khattab2020colbert, gunther2025jina}:
\begin{equation}
s(Q,c)=\sum_{u\in Q}\max_{w\in V_c}\langle u,w\rangle .
\end{equation}
Because text and image entries share one embedding space, every question scores against both views, so a single ranked list can contain text-view and image-view entries for different pages. Each of $q$, $v$, and $e$ retrieves its top-$10$ entries under $s(\cdot,\cdot)$, yielding three ranked lists $T$, $I$, and $E$.

\textbf{Late fusion.} The three lists are merged at page level into a single fused ranking. Our default aggregator is Reciprocal Rank Fusion (RRF) \citep{cormack2009reciprocal}: for a page $p$ with rank $r_L(p)$ in list $L$,
\begin{equation}
\mathrm{RRF}(p)=\sum_{L\in\{T,I,E\}}
  \frac{\mathbf{1}[p\in L]}{k+r_L(p)}, \qquad k=60,
\label{eq:rrf}
\end{equation}
and the top $K{=}10$ pages by $\mathrm{RRF}(p)$ form the retrieved set
$\mathcal{R}_K$. RRF needs no score normalisation and prevents any single list from dominating. Because the best way to combine lists depends on how reliable each one is, we also evaluate three alternatives (Borda count, mean of z-normalised scores, and mean of normalised ranks) in Section~\ref{sec:fusion_strategy}.

\textbf{Answer generation.} The VLM reader is given the question and anchor image $(q,v)$ together with the fused context, and produces the answer $a$. Because fusion merges the lists at the page level, each page in $\mathcal{R}_K$ contributes a single view, i.e., the page text or the page image, whichever entry ranked highest for that page. Hence, the context is a mix of page-text passages and page images. 

\section{AutoQA Benchmark}

AutoQA is a multimodal cross-document question-answering benchmark for the automotive domain. It is built by an automated multi-stage pipeline, with a model-based filtering step after each stage; Appendix~\ref{app:autoqa} describes the pipeline in more detail.

\textbf{Document-pair construction.} We start with 200 automotive workshop manuals rendered into page images. A vision-language model (Qwen3-VL-32B) produces a structured summary of each manual; the summaries are embedded with Jina-v4 and clustered with UMAP \citep{mcinnes2018umap} and HDBSCAN \citep{mcinnes2017hdbscan}. Within each cluster, candidate document pairs are formed from the top nearest neighbours of every manual by summary-embedding
cosine similarity, weighted by cluster-membership confidence. A VLM (Qwen3-VL-32B) then removes pairs that lack shared technical grounding or cross-document complementarity, leaving 556 pairs.

\textbf{Image grounding.} Rather than reuse figures from the manuals, AutoQA grounds each question in a separately sourced image, creating a deliberate domain gap between question images and document pages. 
From the document summaries we generate mechanic-oriented image-search queries, post them to DuckDuckGo Image Search, and filter the returned web images with a VLM to discard noise and mismatches.

\textbf{Question generation.} For each validated pair and its grounding image, GPT-4o generates cross-document questions from the page images, extracted text, and the image, producing 1{,}449 candidate QA pairs. Each question is labelled as factual, procedural, or diagnostic and is constructed to require evidence from both manuals.

\textbf{Filtering.} GPT-4o scored every candidate on factual grounding, answer accuracy, answer completeness, cross-document dependency, absence of source leakage, and image grounding. Retaining items meeting all criteria at maximum quality reduced the 1{,}449 candidates to 885 final QA pairs across 405 distinct images. A 50-item human validation of the final set (Appendix~\ref{app:human_val}) confirms high answerability (80\%) and reference-answer completeness (86\%), with lower strict cross-document necessity (46\%) and image grounding (52\%).

\section{Evaluation}
\subsection{Datasets}
\label{sec:dataset}
We evaluate on three benchmarks spanning different domains and question types. \textbf{MMLongBench} \citep{ma2024mmlongbench} has 1{,}091 questions over long, multi-page documents; with no anchor figures, it is a text-centric setting. \textbf{M3SciQA} \citep{li2024m3sciqa} has 452 cross-document scientific QA pairs,
each grounded in an anchor figure and hinging on tables, charts, and figures. \textbf{AutoQA} (ours) has 885 cross-document automotive QA pairs; each is grounded in a web-sourced image and needs evidence from two manuals, with question images that are noisy and out-of-domain relative to the document pages.

\subsection{Implementation details}
\label{sec:implementation_details}
We use \textit{jina-embeddings-v4} in multi-vector mode (128-dim/token) by default, with dense mode (2048-dim) as an ablation. Page text is extracted with MinerU \citep{wang2024mineru} and indexed with full-page images in one Qdrant collection (top-10 per signal, RRF $k{=}60$; Section~\ref{sec:methodology}). 
The enhanced question is generated by the VLM reader, \textit{Qwen2.5-VL-7B-Instruct}, run on a single NVIDIA H200 GPU. For a controlled comparison, TrioRAG, and all the baselines use the same Qwen2.5-VL-7B reader and the same number of retrieved pages, so accuracy differences reflect retrieval and fusion rather than the reader or context size.

\subsection{Evaluation metric}
\label{sec:evaluation_metrics}
We report mean accuracy under an LLM-as-judge protocol \citep{zheng2023judging} with GPT-4o-mini. Given the question, ground-truth answer, and prediction, the judge
returns a binary score, counting factually consistent predictions correct, including partial answers with the key information and correct refusals on unanswerable questions. Since AutoQA is built with GPT-family models, we use the same judge throughout for comparability and note this dependence in the Limitations.

\section{Results}
\subsection{Main Results}
\label{sec:main_results}

\begin{table}[t]
  \centering
  \small
  \setlength{\tabcolsep}{0pt}
  \begin{tabular*}{\columnwidth}{@{\extracolsep{\fill}}llccc@{}}
    \hline
    \textbf{Method} & \textbf{Index} & \textbf{MMLB} & \textbf{M3Sci} & \textbf{AutoQA} \\
    \hline
    MMGraphRAG  & D & 26.95          & --             & --             \\
    RAG-Anything & D & 27.22          & 28.76          & 58.19          \\
    \hline
    Naive RAG + Concat & D & 33.73   & 31.42          & 55.63          \\
    Naive RAG + Concat & M & 35.56   & 20.69          & 45.31          \\
    \hline
    TrioRAG (Ours) & D & 33.73          & 51.99          & 58.92          \\
    TrioRAG (Ours) & M & \textbf{35.56} & \textbf{54.86} & \textbf{60.11} \\
    \hline
  \end{tabular*}
  \caption{Accuracy (\%) across benchmarks. Index: D = dense, M = multivec.
  Graph baselines use dense indexing; MMGraphRAG is run on MMLongBench.}
  \label{tab:main_results}
\end{table}

Table~\ref{tab:main_results} compares TrioRAG against two graph-based systems, RAG-Anything \citep{guo2025rag} and MMGraphRAG \citep{wan2025mmgraphrag}, that support only dense retrieval; for fairness, we report TrioRAG in both dense and multi-vector modes. 
TrioRAG is best across all three benchmarks. Furthermore, even Naive RAG + Concat (combining signals before retrieval without RRF) outperforms both graph systems on MMLongBench, suggesting that graph construction alone does not inherently improve retrieval quality on text-centric tasks.

The gap is largest on M3SciQA, where TrioRAG nearly doubles RAG-Anything. As M3SciQA is grounded in figures, tables, and charts, direct image retrieval and VLM-enhanced queries exploit visual evidence that graph-based text extraction discards. Feeding full-page images to the reader amplifies this advantage. On AutoQA, the gap narrows to 1.92pp (McNemar $p{=}0.34$, within noise under GPT-4o-mini; under a different-family judge it widens to $+8.14$, $p{<}10^{-5}$, Table~\ref{tab:crossjudge}), as answers are mostly stated in text and both systems rely on text retrieval; we still edge out the graph baseline while being far cheaper (Table~\ref{tab:efficiency}), the trade-off we target. Separate retrieval followed by fusion also outperforms Naive RAG + Concat. Concatenating text and image tokens weakens MaxSim for both modalities, causing a sharp drop for M3SciQA under multi-vector indexing. On MMLongBench, all variants perform identically as the absence of an anchor image collapses all methods to text-only retrieval.

\textbf{Efficiency.} Table~\ref{tab:efficiency} reports total wall time (indexing + inference) on a single H200, with identical settings across methods. Our pipeline completes in under 6 hours in all settings, achieving a 12$\times$ speedup on AutoQA, 20$\times$ on M3SciQA, and 65$\times$ on MMLongBench. The gains come from removing graph construction: both baselines run repeated LLM passes for entity/relation extraction and cross-modal alignment to index the documents, a heavy per-document cost. 
TrioRAG does a single embedding pass at indexing and plain retrieve-and-read at query time, without graph traversal. It is also faster at inference than RAG-Anything. MMGraphRAG has lower query-time latency on MMLongBench but pays for costly graph construction, resulting in higher total runtime while still underperforming in accuracy (Table~\ref{tab:main_results}).

\begin{table}[ht]
  \centering
  \small
  \setlength{\tabcolsep}{4pt}
  \begin{tabular*}{\columnwidth}{@{\extracolsep{\fill}}llrr@{}}
    \hline
    \textbf{Benchmark} & \textbf{Method} & \textbf{Infer (h)} & \textbf{Total (h)}  \\
    \hline
    MMLongBench & MMGraphRAG  & \textbf{1.24} & 66.21  \\
                & RAG-Anything & 3.97 &244.77 \\
                & TrioRAG (Ours) & 1.74 &\textbf{3.75} \\
    \hline
    M3SciQA     & RAG-Anything & 2.40 & 119.53 \\
                & TrioRAG (Ours) & \textbf{1.48} & \textbf{5.93} \\
    \hline
    AutoQA      & RAG-Anything & 4.32 & 62.36 \\
                & TrioRAG (Ours) & \textbf{2.39} & \textbf{5.31} \\
    \hline
  \end{tabular*}
  \caption{Total time (indexing + inference) on a single H200. Inference time refers to the query-time cost, excluding one-time indexing. }
  \label{tab:efficiency}
\end{table}

\begin{table}[ht]
  \centering
  \small
  \setlength{\tabcolsep}{4pt}
  \begin{tabular*}{\columnwidth}{@{\extracolsep{\fill}}lccc@{}}
    \hline
    \textbf{Judge} & \textbf{TrioRAG} & \textbf{RAG-Anything} & \textbf{McNemar $p$} \\
    \hline
    GPT-4o-mini & 60.11 & 58.19 & 0.34 \\
    Opus 4.8    & 38.42 & 30.28 & $<10^{-5}$ \\
    \hline
  \end{tabular*}
  \caption{Cross-family judge accuracy (\%) on 885 AutoQA items. The system ranking is stable and the gap becomes significant under the different-family judge.}
  \label{tab:crossjudge}
\end{table}

\subsection{Ablations}
\label{sec:ablations}

\subsubsection{Indexing} \label{sec:abl-indexing} Multi-vector indexing outperforms dense on all three benchmarks, most on M3SciQA (Table~\ref{tab:main_results}), where answers are grounded in visually distinctive figures. MaxSim's patch-level matching surfaces visually relevant pages that dense pooled embeddings miss, giving the reader richer visual context. We use multi-vector throughout.

\subsubsection{Input representation} 
We compare three reader contexts (Table~\ref{tab:input_rep}):
full-page images; extracted text with cropped figures; and extracted text with full-page images. 
Full-page images preserve layout, charts, and table structure that text extraction and cropping discard, and win on M3SciQA.
Adding extracted text gives a further gain where answers are often stated in text (MMLongBench, AutoQA) but slightly hurts on purely figure-based M3SciQA. As AutoQA is our target setting, we adopt it as the default.

\begin{table}[t]
  \centering
  \small
  \begin{tabular*}{\columnwidth}{@{\extracolsep{\fill}}lccc@{}}
    \hline
    \textbf{Input Representation} & \textbf{MMLB} & \textbf{M3Sci} & \textbf{AutoQA} \\
    \hline
    Doc Page Image            & 34.74          & \textbf{55.75} & 59.48          \\
    Text + Cropped Image      & 31.53          & 51.42          & 57.95          \\
    Text + Doc Page Image     & \textbf{35.56} & 54.86          & \textbf{60.11} \\
    \hline
  \end{tabular*}
  \caption{Accuracy (\%) by reader input representation (MMLB = MMLongBench, M3Sci = M3SciQA).
  We compare three reader contexts: full document-page images; extracted text with cropped figures;
  and extracted text with full-page images.}
  \label{tab:input_rep}
\end{table}

\subsubsection{Question strategy} Table~\ref{tab:question_strategy} compares the three signals under RRF. On M3SciQA, where the anchor figures originate from source documents and retrieves reliably, adding the enhanced question helps (E+I over T+I by 1.99pp, and
the full triple a further 0.44pp). On AutoQA, T+I beats E+I since web-sourced question images mislead the VLM, making $e$ noisy, while text stays reliable. T+I+E is best on both and is our default.

\begin{table}[ht]
  \centering
  \small
  \begin{tabular*}{\columnwidth}{l@{\extracolsep{\fill}}cc}
    \hline
    \textbf{Question Strategy} & \textbf{M3SciQA} & \textbf{AutoQA} \\
    \hline
    T + I       & 52.43          & 59.77          \\
    E + I       & 54.42          & 58.98          \\
    T + I + E   & \textbf{54.86} & \textbf{60.11} \\
    \hline
  \end{tabular*}
  \caption{Accuracy (\%) by Query strategy. T = text, I = image, E = enhanced
  query . T+I+E signal is best.}
  \label{tab:question_strategy}
\end{table}

\subsubsection{Retrieval Reliability and Signal Complementarity}
\label{sec:fusion_strategy}

We measure retrieval at the document level. A signal succeeds on a question if its top-10 contains a page from a reference document (for M3SciQA, the reference paper must be found; the anchor paper is given as the query). We report Recall@10 for each signal and their unions in Table~\ref{tab:retrieval_regime}.

\textbf{Image retrieval works only when the query image is in the corpus.} On M3SciQA, the query image is a figure taken directly from a corpus document. Because of this, the image signal retrieves that source document almost perfectly at 99.6\%. However, this near-duplicate self-matching is deceptive because the actual reference paper is recovered only 46.7\% of the time. On AutoQA, this apparent advantage disappears. AutoQA uses web-sourced images with no counterpart in the corpus, causing image recall to collapse to 19.3\%. This sharp contrast highlights the core problem AutoQA is designed to expose: realistic, out-of-corpus query images defeat visual retrieval. Text and enhanced text signals, by contrast, remain strong and reliable, achieving 81.6\% and 78.5\% recall on AutoQA.

\textbf{Text-derived signals dominate but remain complementary.} Because different query signals fail on different questions, their union outperforms any single signal (87.1\% recall on AutoQA and 97.8\% on M3SciQA). However, these gains come almost entirely from text and enhanced signals. T$\cup$E alone reaches 97.3\% on M3SciQA and 86.6\% on AutoQA, and adding the image signal contributes little beyond that. This pattern provides an empirical basis for retrieving multiple text-derived query signals rather than relying on a single representation. Ultimately, these results foreshadow the limited standalone value of image retrieval when handling realistic queries.

\begin{table}[ht]
  \centering
  \small
  \begin{tabular*}{\columnwidth}{l@{\extracolsep{\fill}}cc}
    \hline
    \textbf{Recall@10} & \textbf{M3SciQA} & \textbf{AutoQA} \\
    \hline
    \multicolumn{3}{l}{\textit{Per signal}} \\
    \quad T        & 95.4 & 81.6 \\
    \quad I       & 46.7 & \textbf{19.3} \\
    \quad E    & 91.2 & 78.5 \\
    \hline
    \multicolumn{3}{l}{\textit{Combined}} \\
    \quad T $\cup$ E      & 97.3 & 86.6 \\
    \quad T $\cup$ I $\cup$ E & \textbf{97.8} & \textbf{87.1} \\
    \hline
    \multicolumn{3}{l}{\textit{Image-Miss subset only}} \\
    \quad T               & --   & 78.9 \\
    \quad T $\cup$ E      & --   & \textbf{84.0} \\
    \hline
  \end{tabular*}
  \caption{Document-level Recall@10 (\%). Image retrieval collapses on AutoQA's
  out-of-corpus query images; the multi-signal union recovers most of
  the loss.}
  \label{tab:retrieval_regime}
\end{table}

\textbf{The enhanced query rescues retrieval when the image signal fails.} The clearest evidence for the enhanced query is on the questions where image retrieval fails entirely. On the 714 AutoQA questions in this regime (80.7\% of the benchmark), adding the enhanced query to text raises reference recall from 78.9\% to 84.0\%. The enhanced query recovers 37 questions that text alone misses, contributing genuinely complementary evidence rather than redundantly re-ranking the same documents text already finds.

\textbf{Detailed retrieval metrics.} Table~\ref{tab:app_retrieval_m3sci} and Table~\ref{tab:app_retrieval_autoqa} report per-signal and fused document-level metrics (R@1, R@5, R@10, MRR@10, nDCG@10) with 95\% bootstrap confidence intervals (10k resamples). We omit MMLongBench, which has no anchor image: $I$ and $E$ are undefined and all aggregators collapse to the text signal (R@10 83.0\%). Notably, M3SciQA reference retrieval is largely a text task: text alone already reaches 95.4 R@10, leaving little headroom for the other signals, whereas on AutoQA the text signal is lower (81.6) and the enhanced query adds complementary coverage.

\begin{table}[t]
  \centering
  \small
  \setlength{\tabcolsep}{3pt}
  \begin{tabular*}{\columnwidth}{l@{\extracolsep{\fill}}ccccc}
    \hline
    \textbf{Run} & \textbf{R@1} & \textbf{R@5} & \textbf{R@10} & \textbf{MRR@10} & \textbf{nDCG@10} \\
    \hline
    T          & 69.0 & 94.5 & 95.4 & 80.9 & 84.6 \\
    I          & 2.2  & 46.7 & 46.7 & 20.6 & 27.2 \\
    E          & 60.6 & 91.2 & 91.2 & 74.9 & 79.1 \\
    RRF        & 48.0 & 96.9 & 96.9 & 70.5 & 77.3 \\
    Borda      & 49.6 & 97.1 & 97.8 & 71.8 & 78.5 \\
    Score-mean & 33.6 & 96.7 & 97.6 & 63.3 & 72.1 \\
    Rank-mean  & 38.7 & 96.0 & 97.6 & 64.9 & 73.3 \\
    \hline
  \end{tabular*}
  \caption{M3SciQA ($n{=}452$) document-level retrieval metrics (\%).}
  \label{tab:app_retrieval_m3sci}
\end{table}

\begin{table}[t]
  \centering
  \small
  \setlength{\tabcolsep}{3pt}
  \begin{tabular*}{\columnwidth}{l@{\extracolsep{\fill}}ccccc}
    \hline
    \textbf{Run} & \textbf{R@1} & \textbf{R@5} & \textbf{R@10} & \textbf{MRR@10} & \textbf{nDCG@10} \\
    \hline
    T          & 31.4 & 74.2 & 81.6 & 48.8 & 45.5 \\
    I          & 4.4  & 14.7 & 19.3 & 8.6  & 8.6  \\
    E          & 31.0 & 71.9 & 78.5 & 47.3 & 44.0 \\
    RRF        & 33.1 & 72.3 & 79.0 & 49.2 & 44.3 \\
    Borda      & 34.0 & 71.4 & 82.1 & 49.9 & 46.1 \\
    Score-mean & 11.4 & 61.4 & 81.7 & 31.8 & 35.9 \\
    Rank-mean  & 11.3 & 59.8 & 80.8 & 31.1 & 35.1 \\
    \hline
  \end{tabular*}
  \caption{AutoQA ($n{=}885$) document-level retrieval metrics (\%).}
  \label{tab:app_retrieval_autoqa}
\end{table}

\textbf{Fusion method has little effect.} Given reliable signals, the choice among rank aggregators matters far less than the signals themselves. Across four unsupervised aggregators (Borda count, score mean, rank mean, RRF), accuracy differences are not statistically significant on either benchmark (Table~\ref{tab:overall_fusion_accuracy}; paired tests, Borda vs.\ RRF: $p\approx0.31$ on M3SciQA, $p\approx0.45$ on AutoQA). RRF is also insensitive to the rank constant on AutoQA (accuracy $60.11, 60.0, 59.77$ at $k{=}60, 15, 20$). We therefore adopt RRF, the simplest option requiring no score normalisation, as the default.

\textbf{Fusion shifts early ranks, not coverage.} The detailed metrics (Tables~\ref{tab:app_retrieval_m3sci}--\ref{tab:app_retrieval_autoqa}) refine this picture. Across the four aggregators, coverage barely moves (AutoQA R@10 spans only 3.1pp), whereas early ranks vary widely (MRR@10 spans 18.8pp). On M3SciQA, fusing the three signals lowers text-only R@1 from 69.0 to 48.0 while raising R@10 from 95.4 to 96.9: fusion trades top-rank precision for broader coverage. Because the reader consumes the full top-10, coverage rather than early-rank order is what drives downstream accuracy, which is why the aggregators are statistically indistinguishable in accuracy (Table~\ref{tab:overall_fusion_accuracy}).

\textbf{Retrieval success drives accuracy.} To connect retrieval quality directly to answers, we split TrioRAG's accuracy (with the fixed Qwen2.5-VL-7B reader) by whether the gold document was retrieved in the top-10 (Table~\ref{tab:cond_retrieval}). On every benchmark and under both judges, retrieving the gold document raises accuracy substantially: by $+15.5$pp (GPT-4o-mini) and $+22.8$pp (Opus~4.8) on AutoQA, $+38.9$pp on M3SciQA, and $+25.0$pp on MMLongBench. Retrieval outcome thus moves accuracy far more than the choice of aggregator (a $1.24$pp spread on AutoQA, Table~\ref{tab:overall_fusion_accuracy}). The analysis is observational (accuracy tracks rather than causes retrieval success), but it reinforces that retrieval reliability governs downstream accuracy.

\begin{table}[t]
  \centering
  \small
  \setlength{\tabcolsep}{4pt}
  \begin{tabular*}{\columnwidth}{@{\extracolsep{\fill}}llccc@{}}
    \hline
    \textbf{Bench} & \textbf{Judge} & \textbf{Acc$_{\text{ret}}$} & \textbf{Acc$_{\text{miss}}$} \\
    \hline
    AutoQA  & GPT-4o-mini & 63.4 & 47.8 \\
    AutoQA  & Opus 4.8    & 43.2 & 20.4 \\
    M3SciQA & GPT-4o-mini & 53.2 & 14.3 \\
    MMLB    & GPT-4o-mini & 38.0 & 13.0 \\
    \hline
  \end{tabular*}
  \caption{TrioRAG accuracy (\%) with the fixed Qwen2.5-VL-7B reader, split by whether the gold document was in the retrieved top-10 (Acc$_{\text{ret}}$) or missed (Acc$_{\text{miss}}$); the \textbf{Judge} column is the LLM-as-judge that scored correctness (AutoQA shown under both model families).}
  \label{tab:cond_retrieval}
\end{table}

\begin{table}[t]
  \centering
  \small
  \renewcommand{\arraystretch}{1.15}
  \begin{tabular*}{\columnwidth}{l@{\extracolsep{\fill}}cc}
    \hline
    \textbf{Fusion Method} & \textbf{M3SciQA} & \textbf{AutoQA} \\
    \hline
    Borda      & \textbf{57.08} \scriptsize(52.7--61.7) & 59.66 \scriptsize(56.4--62.9) \\
    Score Mean & 56.86 \scriptsize(52.2--61.5)          & 58.87 \scriptsize(55.6--62.2) \\
    Rank Mean  & 55.31 \scriptsize(50.7--60.0)          & 59.55 \scriptsize(56.3--62.9) \\
    RRF        & 54.86 \scriptsize(50.4--59.5)          & \textbf{60.11} \scriptsize(56.8--63.3) \\
    \hline
  \end{tabular*}
  \caption{Accuracy (\%) by fusion method with 95\% bootstrap CIs (10k resamples).}
  \label{tab:overall_fusion_accuracy}
\end{table}

\section{Conclusion}
In this paper, we propose TrioRAG, a graph-free, multi-signal pipeline that matches or exceeds the graph-based systems we evaluate.
We evaluate it on three benchmarks, with per-query inference 1.6 to 2.3$\times$ faster (Table~\ref{tab:efficiency}). Our central finding is that retrieval reliability matters more for downstream accuracy than the choice of fusion method: text-derived signals keep retrieval robust where image retrieval collapses on out-of-corpus query images. Because indexing is a single embedding pass, the graph-free design also keeps the corpus cheap to update: adding a document requires only re-embedding the changed pages, with no entity re-extraction, cross-modal re-alignment, or graph rebuilding. These results suggest that, for multimodal cross-document QA, retrieval-signal quality matters more than pipeline complexity. Promising directions include stronger visual signals for out-of-corpus queries and adaptive, per-question weighting of the text, image, and enhanced signals.
\section*{Limitations}
\label{sec:limitations}
\textbf{Model-curated benchmark.} AutoQA is generated and filtered by vision-language models, so reported accuracies reflect a model-curated reference rather than a human gold standard. To gauge quality, we ran a 50-item human validation (Appendix~\ref{app:human_val}): answerability (80\%) and reference-answer completeness (86\%) are high, while strict cross-document necessity (46\%) and image grounding (52\%) are lower. These measured rates should be kept in mind when reading AutoQA results; a larger study with inter-annotator agreement is left to future work.

\textbf{Judge and construction share a model family.} Answers are produced by Qwen2.5-VL-7B and judged by GPT-4o-mini, so there is no answerer--judge circularity; however, AutoQA is generated by GPT-4o and judged by GPT-4o-mini, both in the GPT family, which may inflate its absolute scores. To bound this, we re-judged all AutoQA items with a different-family judge (Claude Opus~4.8, Table~\ref{tab:crossjudge}): absolute scores fall for all systems, but the ranking is unchanged and the TrioRAG--RAG-Anything gap grows and becomes significant, consistent with GPT-family inflation rather than a reversal. We keep a single judge across datasets for comparability.

\textbf{Fusion choice.} The four aggregators we compare do not differ significantly in accuracy (Table~\ref{tab:overall_fusion_accuracy}). We default to RRF because it needs no score normalization and keeps any single signal from dominating the fused ranking, a useful property given the heterogeneous reliability of our three signals; we do not claim it is optimal.

\textbf{Single component choices.} All methods (TrioRAG and the graph-based baselines) share the same reader (Qwen2.5-VL-7B) and encoder (Jina-v4), so these fixed components do not explain the accuracy differences, which stem from retrieval and fusion design. Absolute accuracies would change with a stronger reader; whether the relative gains hold across reader and encoder families is left to future work.

% \textbf{Deployment characterization.} We report index footprint and per-query
% retrieval cost (Appendix~\ref{sec:index_footprint}) but do not measure sustained
% throughput under concurrent serving load. The update path is nonetheless cheaper
% than in graph systems: adding a document requires a single embedding pass and a
% vector insertion, with no entity re-extraction, cross-modal re-alignment, or
% graph rebuilding.

\section*{Ethics Statement}
\label{sec:ethics}

\textbf{Source images and licensing.} AutoQA grounds each question in a web image (via DuckDuckGo Image Search) that we did not author and that may carry copyright. Rather than redistribute the files, we release each image's source URL and a SHA-256 content hash, so users fetch images from their original sources and can verify they match ours even if some links later break.

\textbf{Source documents and release.} The workshop manuals are proprietary service documents, so we do not redistribute them. Image URLs and content hashes carry no proprietary content and will be released; release of the QA items and document references is subject to institutional review, and we will release the maximally reproducible subset permitted.

\textbf{Model-curated construction.} AutoQA is built with Qwen3-VL-32B (summaries, pairing, image search and verification) and GPT-4o (QA generation and verification); as noted in the Limitations, it is a model-curated testbed rather than a human-validated gold standard and its absolute accuracies should be read accordingly.

\textbf{Intended use and risks.} AutoQA targets retrieval and reasoning over automotive technical documentation. It is a research benchmark and is not intended to provide safety-critical repair guidance; answers produced by systems evaluated on it should not be relied upon for real vehicle servicing without expert verification.

\textbf{AI assistance.} The authors used AI-based tools to assist with drafting and editing the manuscript. All scientific claims, experimental design, and analysis are the authors' own, and the authors take full responsibility for the content.

\section*{Acknowledgments}
The authors thank the International Max Planck Research School for Intelligent Systems (IMPRS-IS) for supporting Yuqicheng Zhu and Hongkuan Zhou.

\bibliography{custom}

\appendix
\section{Appendix}
\label{sec:appendix}

\begin{figure*}[t]
  \centering
  \includegraphics[width=0.95\textwidth]{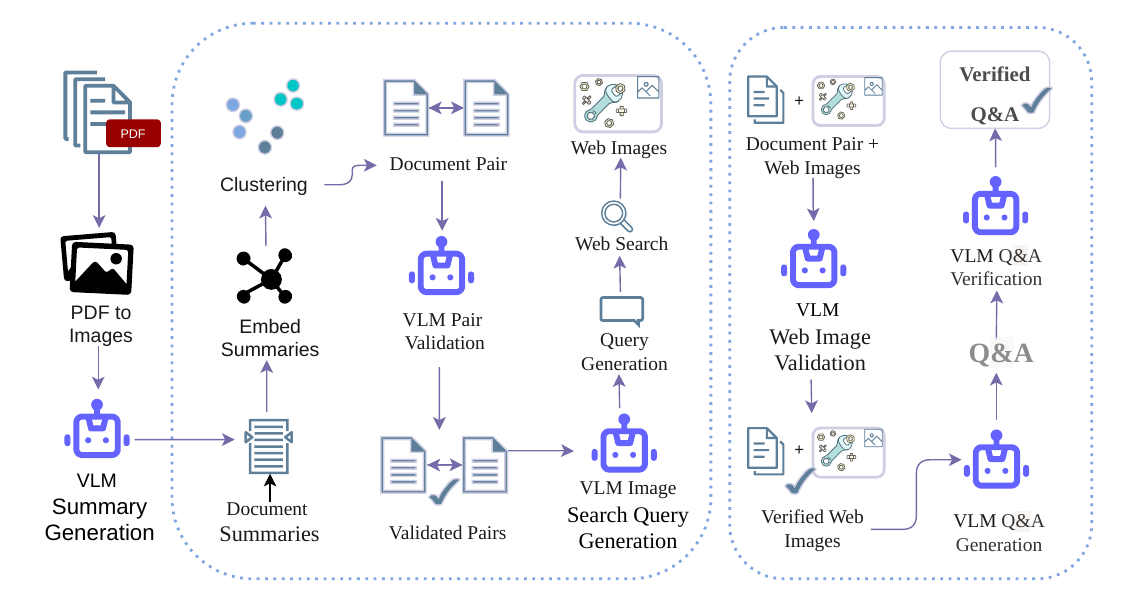}
  \caption{The AutoQA construction pipeline: manuals are summarised and clustered
  to form document pairs, each pair is grounded in a web-sourced image, and GPT-4o
  generates and filters cross-document QA pairs.}
  \label{fig:autoqa_pipeline}
\end{figure*}

\subsection{AutoQA construction details}
\label{app:autoqa}

AutoQA is created through an automated multi-stage pipeline, with a model-based filtering step applied after each stage (Figure~\ref{fig:autoqa_pipeline}).

\textbf{Document selection.} We begin with 200 workshop manuals selected from a larger HTML corpus. Documents are ranked using a composite score that considers length, vocabulary richness, figure and table density, heading depth, cross-references, and automotive-term density. Near-duplicate documents are removed before selection.

\textbf{Clustering and pairing.} Each manual is first summarized using Qwen3-VL-32B. The summaries are embedded with Jina-v4, reduced using UMAP ($10$ components, $15$ neighbours, minimum distance $0.1$, cosine metric), and clustered with HDBSCAN (minimum cluster size $3$, minimum samples $2$, Excess of Mass (EOM) selection). This process produces $17$ clusters, while low-confidence assignments are treated as noise. Within each cluster, candidate document pairs are formed from the top five cosine neighbours of every document. Pairs are scored as $0.7\cdot\mathrm{cos}(i,j) + 0.3\cdot(\mathrm{prob}_i\cdot\mathrm{prob}_j)$, resulting in $591$ candidate pairs. Qwen3-VL-32B then performs a lightweight screening step to verify the presence of a shared technical topic and a meaningful cross-document dependency. This removes $35$ pairs, leaving $556$ for the next stage.

\textbf{Image grounding.} For each document pair, Qwen3-VL-32B generates three mechanic-style image search queries. These queries are submitted to DuckDuckGo Image Search, and the returned photographs are filtered by Qwen3-VL-32B to ensure they depict relevant automotive components, are real-world images, and meet a minimum quality threshold (confidence $>0.8$).

\textbf{Question generation and filtering.} GPT-4o generates up to three cross-document question-answer pairs for each document pair using a temperature of $0.3$. Each question is grounded in a verified image and in $20$ sampled pages from each document, producing $1{,}449$ candidate QA pairs. GPT-4o then evaluates each candidate on factual grounding, answer accuracy, answer completeness, cross-document dependency, absence of source leakage, and image grounding. Only candidates that satisfy every criterion and receive the highest overall quality rating are retained. Cross-document dependency and overall quality are the most restrictive criteria, with pass rates of $71.8\%$ and $72.0\%$, respectively; the remaining criteria rarely eliminate candidates.

The final dataset contains $885$ QA pairs grounded in $405$ distinct web-sourced images, with multiple questions sharing the same image when applicable. Questions are categorized as procedural ($340$, $38.4\%$), diagnostic ($292$, $33.0\%$), or factual ($253$, $28.6\%$).

\subsection{Human validation of AutoQA}
\label{app:human_val}
To provide an independent, non-model check of benchmark quality, we drew a uniform random sample of $50$ final AutoQA items and had a human annotator label each for four properties (Table~\ref{tab:human_val}). Answerability ($80\%$) and reference-answer completeness ($86\%$) are high, whereas strict cross-document necessity ($46\%$) and correct image grounding ($52\%$) are lower. The human cross-document-necessity rate ($46\%$) is below the $71.8\%$ reported by our GPT-4o construction filter (Appendix~\ref{app:autoqa}); within the final $885$ items this field is always true by construction, so the filter cannot itself validate it. Part of the gap also reflects a stricter human criterion (both manuals must be \emph{necessary}, rather than merely combining information from both). We therefore report these measured rates rather than assuming uniform benchmark quality; a larger study with inter-annotator agreement is a natural next step.

\begin{table}[ht]
  \centering
  \small
  \begin{tabular*}{\columnwidth}{l@{\extracolsep{\fill}}c}
    \hline
    \textbf{Property} & \textbf{Human pass rate} \\
    \hline
    Answerable                & 40/50 (80\%) \\
    Requires both manuals     & 23/50 (46\%) \\
    Correctly image-grounded  & 26/50 (52\%) \\
    Reference answer complete  & 43/50 (86\%) \\
    \hline
  \end{tabular*}
  \caption{Human validation on a uniform random sample of 50 AutoQA items.}
  \label{tab:human_val}
\end{table}

% =====================================================================
% Appendix: Qualitative AutoQA examples  (Examples: id 852, 387, 1072)
% Package-free. Verdict marks use \textsc{}; swap for \checkmark /
% \ding{55} if amssymb / pifont are already loaded.
% Anchor figures are SYNTHETIC illustrations of the image type, not the
% real web-sourced anchors (consistent with the Ethics Statement, which
% releases only URLs + SHA-256 hashes). Disclose the AI-generated
% figures in the Ethics Statement.
% Image files expected at latex/images/generated_eg{1,2,3}.png
% =====================================================================

\subsection{Qualitative Examples}
\label{app:examples}

We present three AutoQA items, each with the TrioRAG prediction
(multi-vector index, RRF fusion, Qwen2.5-VL-7B reader) and the
GPT-4o-mini judge verdict. Every question is grounded in a web-sourced
anchor photo and needs evidence from two manuals. The examples span the three
question types and include one failure that illustrates a characteristic cross-document error.  Questions, reference answers, and judge rationales are verbatim; predictions are abridged, with $[\ldots]$ marking omitted text. The anchor figures are synthetic illustrations of each image type, not the actual benchmark images, which we release only
as source URLs and content hashes (see Ethics Statement).

\begin{figure}[t]
    \centering
    \includegraphics[width=\columnwidth, trim=10 180 0 180, clip]
        {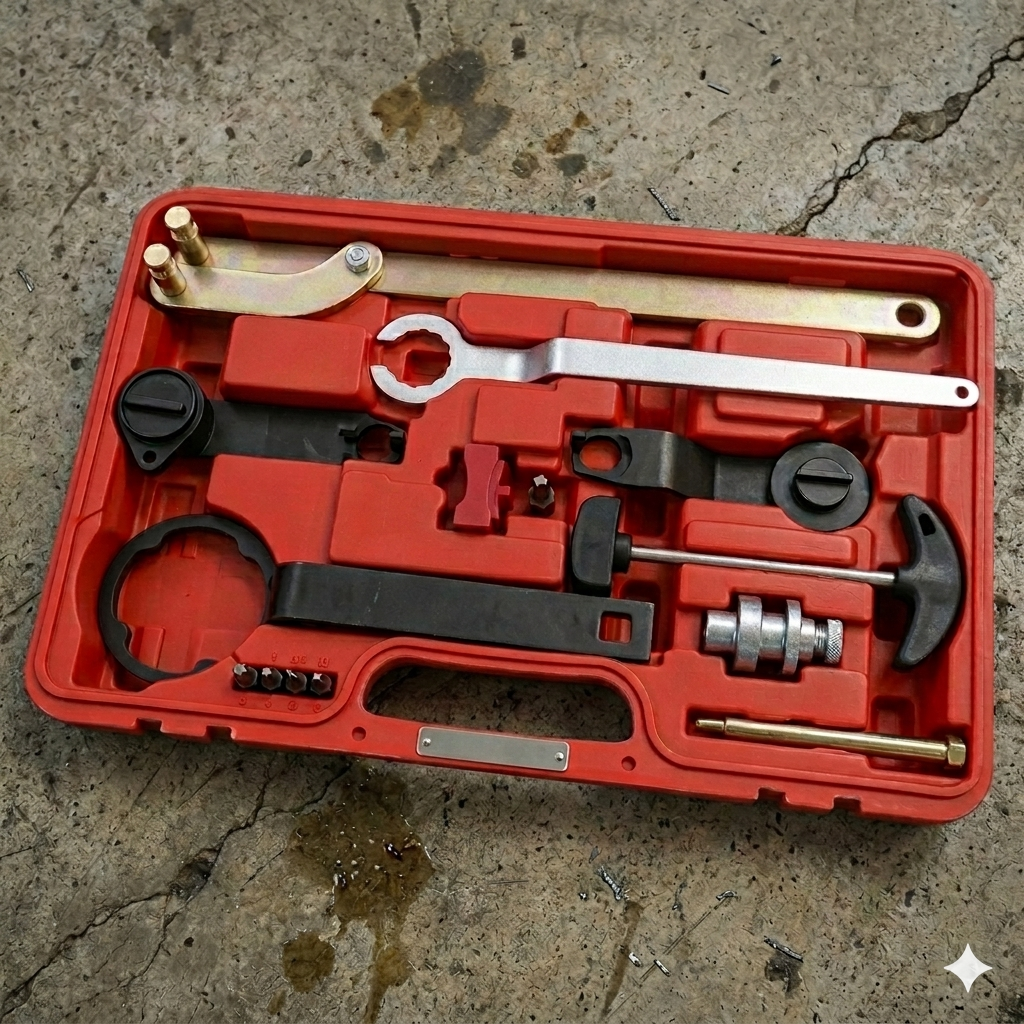}
    \caption{Synthetic illustration of the anchor image used in Example~1: a
    camshaft counter-hold toolkit. This is not the actual benchmark image.}
    \label{fig:eg1}
\end{figure}

% =====================================================================
\paragraph{Example 1: Procedural (Judged correct).}
The anchor photo (Fig.~\ref{fig:eg1}) shows a counter-hold toolkit.

\textbf{Question.} In the photo, I see a counter-hold tool similar to T10554. How do I use it to remove the camshaft adjuster and ensure proper alignment during
installation?

\textbf{Reference answer.} To remove the camshaft adjuster, fit the counter-hold tool T10554 with
T10172A to the camshaft pulley, ensuring the pins are properly inserted
into the holes. Hold the inlet camshaft in position using these tools
while loosening the securing bolt of the camshaft adjuster. During
installation, ensure the guide sleeve is inserted in the camshaft and
renew the bolt for the camshaft adjuster, screwing it in by hand. The
camshaft pulleys should be able to turn without rocking.

\textbf{TrioRAG prediction (abridged).} Position counter-hold tool T10554 so its pins align with the holes in
the camshaft adjuster and secure it with T10172A. Loosen the securing
bolt one turn without fully removing it, then withdraw the adjuster from
the camshaft. $[\ldots]$ To reinstall, refit T10554/T10172A to hold the
camshaft, realign the adjuster, and tighten the securing bolt without
over-tightening. $[\ldots]$ Finally, confirm the adjuster is correctly
aligned with the camshaft.

\textbf{Judge (correct).}
``The generated answer provides a correct and detailed procedure for
using the counter-hold tool T10554 to remove and reinstall the camshaft
adjuster, ensuring proper alignment, which matches the expected answer.''

\begin{figure}[t]
    \centering
    \includegraphics[width=\columnwidth, trim=10 130 0 0, clip]
        {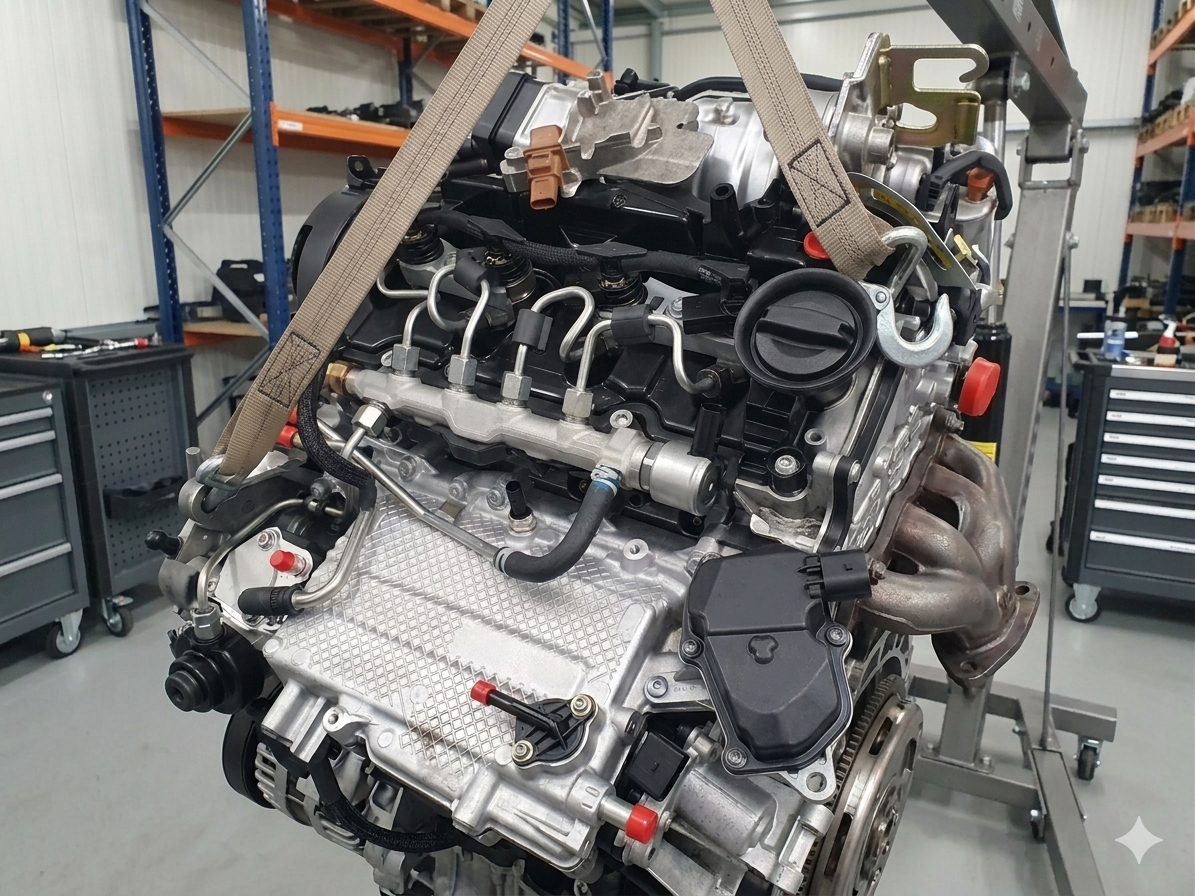}
    \caption{Synthetic illustration of the anchor image used in Example~2: a
    transverse four-cylinder engine showing the fuel rail, injector
    connectors, and engine-control-unit connector.}
    \label{fig:eg2}
\end{figure}

% =====================================================================
\paragraph{Example 2: Factual (judged incorrect).}
The anchor photo (Fig.~\ref{fig:eg2}) shows an
engine-control-unit connector with several fuel-injector leads.

\textbf{Question.}
In the photo, I see a 91-pin connector and several fuel injector
connections. How do these components interact in the 2.0L diesel engine and the 1.6L petrol engine?

\textbf{Reference answer.}
In the 2.0L diesel engine, the 91-pin connector is linked to the
engine/motor control unit (J623), which manages the injectors for
cylinders 1 to 4 (N30, N31, N32, N33). For the 1.6L petrol engine, the
94-pin connector connects to the engine/motor control unit (J623),
controlling the injectors similarly but with different wiring
configurations (T2ln, T2lo, T2lp, T2lq). Both engines use their
respective connectors to facilitate precise fuel injection timing and
control.

\textbf{TrioRAG prediction (abridged).}
The 91-pin connector is the interface between the ECU and the engine's
sensors and actuators, and it controls the fuel injectors in both the
2.0L diesel and the 1.6L petrol engine. $[\ldots]$ In both engines the
ECU sends signals through the 91-pin connector to open and close the
injectors, with only minor differences due to engine design.

\textbf{Judge (incorrect).}
``The generated answer incorrectly refers to a 91-pin connector for both engines instead of specifying the correct connectors for each engine; it also omits exact details about how the fuel injectors are linked to the different engine configurations as specified in the expected answer.''

The model substitutes generic ECU boilerplate for the petrol engine's 94-pin connector and the cylinder-specific injector labels, a failure that requires jointly grounding both manuals rather than either alone.

\begin{figure}[t]
    \centering
    \includegraphics[width=\columnwidth, trim=10 130 0 0, clip]
        {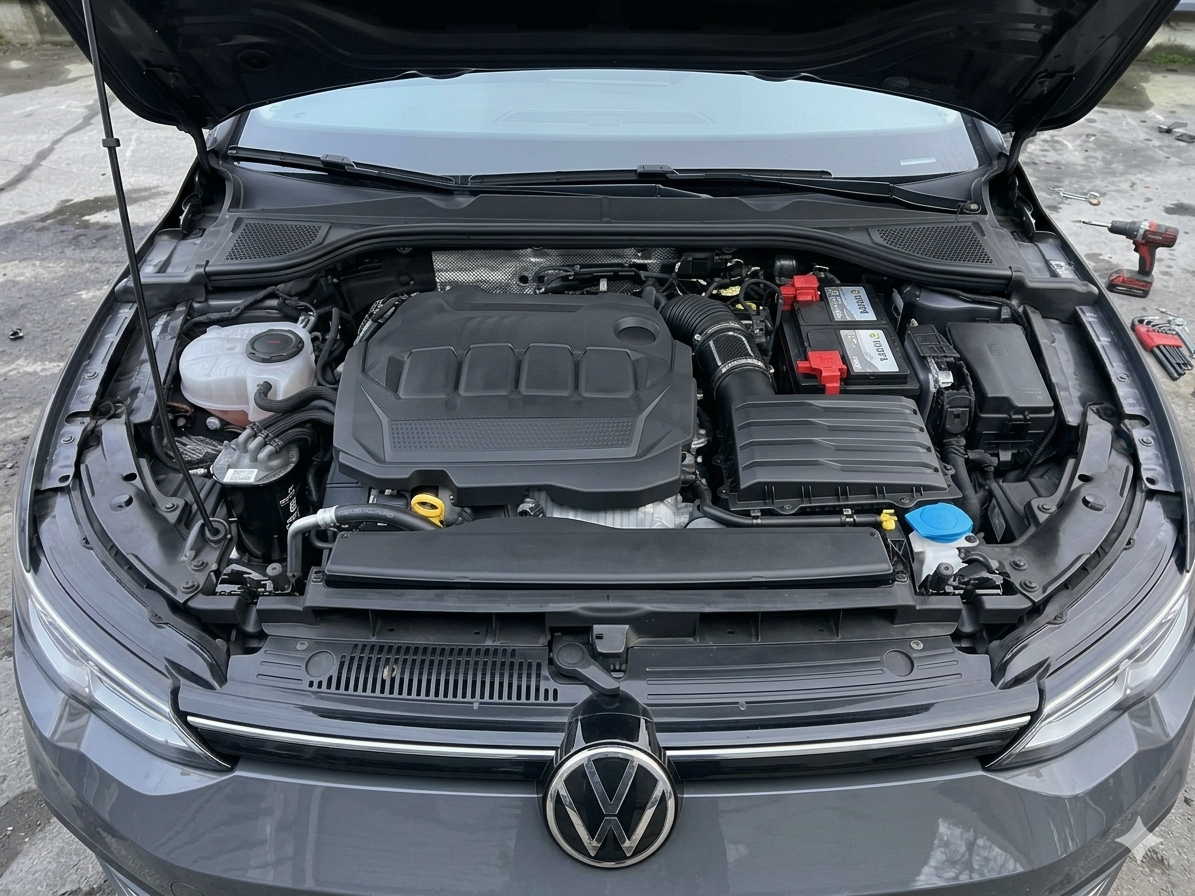}
    \caption{Synthetic illustration of the anchor image
used in Example~3: an
    engine-bay fuse box beside the battery. This is not the actual
    benchmark image.}
    \label{fig:eg3}
\end{figure}

% =====================================================================
\paragraph{Example 3: Diagnostic (judged correct).}
The anchor photo (Fig.~\ref{fig:eg3}) shows an engine-bay fuse
box beside the battery.

\textbf{Question.}
In the photo, I see the fuse holder near the battery. Which fuses should
I check to troubleshoot the gearbox communication issue?

\textbf{Reference answer.}
Check Fuse 12 on Fuse Holder B (SB12) and Fuse 6 on Fuse Holder C (SC6) as they are related to the automatic gearbox control unit. These fuses
are crucial for ensuring proper communication between the gearbox and
other systems. Inspect them for continuity and replace if necessary.

\textbf{TrioRAG prediction (abridged).}
The fuse holder near the battery is Fuse holder~B (SB). The fuse related
to the automatic gearbox control unit (J217) is Fuse~12 (SB12). With the
ignition off, locate holder~B and check whether SB12 is blown, replacing
it if so. $[\ldots]$ If SB12 is intact, check the gearbox relays, the
wiring harnesses to the control unit, and the control unit itself,
consulting the ``Relays on fuse holder BSB'' section for relay
locations.

\textbf{Judge (correct).}
``The generated answer correctly identifies Fuse 12 (SB12) related to
the gearbox control unit and provides troubleshooting steps that align
with the expected answer.''
\textit{(Top-10 fused pages spanning 8 documents.)} The reader recovers
the holder-B gearbox fuse SB12 but not the second relevant fuse SC6 on
holder~C; the judge accepts the partially complete answer, consistent
with the partial-credit protocol in Section~\ref{sec:evaluation_metrics}.

\subsection{Prompts}
\label{app:prompts}

\paragraph{AutoQA QA generation.} GPT-4o receives sampled pages from both documents
and the anchor photo, and is instructed to produce QA pairs. Core instructions below (abridged):
\begin{quote}\small\ttfamily
You are generating a cross-document multimodal QA dataset for automotive
workshop manuals. You are given sampled pages from Document A and Document B
and one real-world mechanic photo.\\[2pt]
Requirements:\\
1. Each question is from a mechanic referencing the photo.\\
2. Each question must mention at least one visible detail from the photo.\\
3. Each pair must require both documents, combining distinct information from each.\\
4. The answer must fully resolve the question with document-specific details.\\
5. Rely on document-specific details only; do not add generic advice.\\
6. Use a mix of factual, procedural, and diagnostic questions.\\
7. Do not mention document names, page numbers, or ``Document A/B''.\\
8. State facts and procedures directly, as an automotive expert would.\\[2pt]
Output a JSON array of pairs, each with question, answer, question\_type, and
reference\_docs.
\end{quote}

\paragraph{Enhanced query (AutoQA).} The VLM reader is prompted with the question and anchor image to produce the enhanced query; prompts for the other benchmarks differ only in domain framing.
\begin{quote}\small\ttfamily
You are given an automotive figure and a question about it. Your task is to
generate a grounded natural language question that integrates both the
textual question and the visual information from the figure.\\[2pt]
Instructions:\\
- Carefully analyze the question and the figure simultaneously.\\
- Identify what parts of the figure are relevant to the question.\\
- Produce a single, well-formed natural language question that is explicitly
grounded in the visual content.\\
- Preserve the original intent of the question.\\
- The output question should be more specific and informative than the
original question alone.\\
- Do NOT answer the question. Do NOT explain. Do NOT invent details.\\[2pt]
Question: \{question\}\\
Grounded question:
\end{quote}

\textbf{Answer judging.} At evaluation, GPT-4o-mini receives the question,
the ground-truth answer, and the prediction, and returns a binary score.
\begin{quote}\small\ttfamily
You are an expert evaluator assessing answers from a multimodal document QA
system. Evaluate whether the generated answer correctly responds to the
question based on the expected answer.\\[2pt]
Question: \{question\}\\
Expected Answer: \{answer\}\\
Generated Answer: \{prediction\}\\[2pt]
Accuracy (0 or 1): Does the generated answer match the factual content of the
expected answer?\\
- 1: Factually correct and aligns with the expected answer.\\
- 0: Factually incorrect or contradicts the expected answer.\\[2pt]
Instructions:\\
- Focus on factual correctness, not writing style or format.\\
- Partial matches are accurate if the correct information is present.\\
- For numerical answers, check if values match or are equivalent.\\
- For list answers, check if all key elements are present.\\
- If the expected answer is ``Not answerable'' and the generated answer
indicates inability to answer, consider it accurate.\\
\end{quote}

\subsection{Index Footprint}
\label{sec:index_footprint}
For AutoQA, the unified text-and-page-image multivector collection stores 7{,}682 page-level points and occupies 4.2~GB on disk
(3.9~GiB), including token-level vectors and payload. Each page is indexed as a single point whose multivector representation is computed in one embedding pass with \textit{jina-embeddings-v4}. 
\end{document}